\documentclass[conference]{IEEEtran}
\ifCLASSINFOpdf
\else
\fi

\usepackage{graphicx}
\usepackage{subcaption}
\usepackage{caption}
\usepackage{multirow}
\usepackage{amsmath}
\usepackage[euler]{textgreek}

\usepackage{fancyhdr}

\begin{document}

\title{
An Improved Approach for Prediction of Parkinson's Disease using Machine Learning Techniques
}

\author{\IEEEauthorblockN{Kamal Nayan Reddy Challa}
\IEEEauthorblockA{
School of Electrical Sciences\\
Computer Science and Engineering\\
Indian Institute of Technology\\
Bhubaneswar, India 751013\\
Email: kc11@iitbbs.ac.in}
\and
\IEEEauthorblockN{Venkata Sasank Pagolu}
\IEEEauthorblockA{
School of Electrical Sciences\\
Computer Science and Engineering\\
Indian Institute of Technology\\
Bhubaneswar, India 751013\\
Email: vp12@iitbbs.ac.in}
\and
\IEEEauthorblockN{Ganapati Panda}
\IEEEauthorblockA{
School of Electrical Sciences\\
Indian Institute of Technology\\
Bhubaneswar, India 751013\\
Email: gpanda@iitbbs.ac.in}
\and
\IEEEauthorblockN{\hspace{20pt}Babita Majhi}
\IEEEauthorblockA{
	\hspace{25pt}Department of Computer Science and IT\\
	\hspace{25pt}G.G Vishwavidyalaya, Central University \\
\hspace{25pt}Bilaspur, India 495009\\
\hspace{25pt}Email: babita.majhi@gmail.com}
}

\maketitle
\thispagestyle{fancy}

\begin{abstract}
Parkinson's disease (PD) is one of the major public health problems in the world. It is a well-known fact that around one million people suffer from Parkinson's disease in the United States whereas the number of people suffering from Parkinson's disease worldwide is around 5 millions. Thus, it is important to predict Parkinson's disease in early stages so that early plan for the necessary treatment can be made. People are mostly familiar with the motor symptoms of Parkinson's disease, however an increasing amount of research is being done to predict the Parkinson's disease from non-motor symptoms that precede the motor ones. If early and reliable prediction is possible then a patient can get a proper treatment at the right time. Non-motor symptoms considered are Rapid Eye Movement (REM) sleep Behaviour Disorder (RBD) and olfactory loss. Developing machine learning models that can help us in predicting the disease can play a vital role in early prediction. In this paper we extend a work which used the non-motor features such as RBD and olfactory loss. Along with this the extended work also uses important biomarkers. In this paper we try to model this classifier using different machine learning models that have not been used before. We developed automated diagnostic models using Multilayer Perceptron, BayesNet, Random Forest and Boosted Logistic Regression.  It has been observed that Boosted Logistic Regression provides the best performance with an impressive accuracy of 97.159 \% and the area under the ROC curve was 98.9\%. Thus, it is concluded that this models can be used for early prediction of Parkinson's disease.
\end{abstract}
\IEEEpeerreviewmaketitle

\section{Introduction}
Parkinson's disease (PD) is a chronic, degenerative neurological disorder. The main cause of Parkinson's disease is actually unknown. However, it has been researched that the combination of environmental and genetic factors play an important role in causing PD [1]. For general understanding the Parkinson's disease is treated as disorder of the central nervous system which is the result of loss of cells from various parts of the brain. These cells also include substantia nigra cells that produce dopamine. Dopamine plays a vital role in the coordination of movement. It acts as a chemical messenger for transmitting signals within the brain. Due to the loss of these cells, patients suffer from movement disorder. 

The symptoms of PD can be classified into two types i.e. non-motor and motor symptoms. Many people are aware of the motor symptoms as they can be visually perceived by human beings. These symptoms are also called as cardinal symptoms, these include resting tremor, slowness of movement (bradykinesia), postural instability (balance problems) and rigidity [2]. It is now established that there exists a time-span in which the non-motor symptoms can be observed. This symptoms are called as “dopamine-non-responsive” symptoms. These symptoms include cognitive impairment, sleep difficulties, loss of sense of smell, constipation, speech and swallowing problems, unexplained pains, drooling, constipation and low blood pressure when standing. It must be noted that none of these non-motor symptoms are decisive, however when these features are used along with other biomarkers from Cerebrospinal Fluid measurement (CSF) and dopamine transporter imaging, they may help us to predict the PD.

In this paper we extend works by Prashant et al [3]. This work takes into consideration the non-motor symptoms and the biomarkers such as cerebrospinal fluid measurements and dopamine transporter imaging. In this paper we follow a similar approach, however we try to use different machine learning algorithms that can help in improving the performance of model and also play a vital role in making in early prediction of PD which in turn will help us to initiate neuroprotective therapies at the right time.

\begin{figure*}[h]
  \centering
  \includegraphics[]{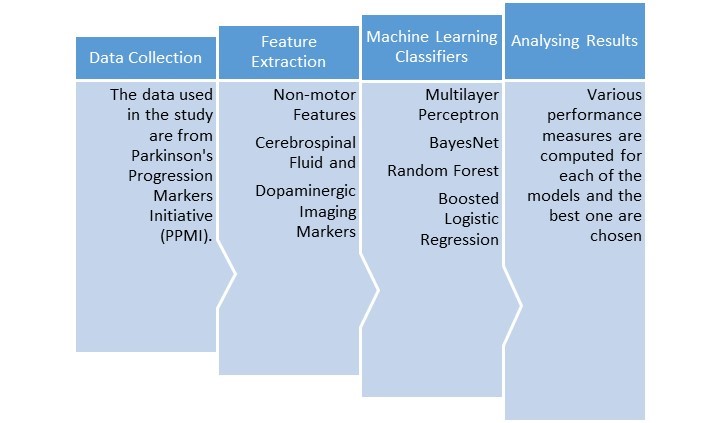}
  \caption{Flow Chart of the proposed analysis}
  \label{fig: Flow Chart of the proposed analysis}
\end{figure*}

The rest of the paper is organized as follows. Section 2 contains the related work. Section 3 contains the flowchart of the analysis carried out and describes about the PPMI database, explanation of different features extracted, statistical analysis of this features, classification and prediction/prognostic model design. Section 4 provides the results and discussion from the experiments carried out. And finally conclusion of the work is provided in Section 5.

\section{Related Research Work}
Different researchers have used different features and data to predict Parkinson's disease. Indira et al. [4] have used biomedical voice of human as the main feature. The authors have developed a model to automatically predict whether a person is suffering from PD by analysing the voice of the patients. They have used fuzzy c-means (FCM) clustering and pattern recognition methods on the dataset and have attained an accuracy of 68.04\%, 75.34\% sensitivity and 45.83\% specificity.  Amit et al. [5] have presented a unique approach of classifying PD patients on the basis of their postural instability and have used L2 norm metric in conjunction with support vector machine.   In [6], the authors have applied University of Pennsylvania 40-item smell identification test (UPSIT-40) and 16-item identification test from Sniffin’s Sticks. This study was conducted on Brazilian population. The authors have applied logistic regression considering each of the above features separately. They observed that the Sniffin’ Sticks gave a specificity of 89.0 \% and a sensitivity of 81.1 \%.  Similarly they found out that the UPSIT-40 specificity was 83.5\% and sensitivity 82.1\%. Prashant et al.  [7] have used olfactory loss feature loss from 40-item UPSIT and sleep behaviour disorder from Rapid eye movement sleep Behaviour Disorder Screening Questionnaire (RBDSQ). Support Vector machine and classification tree methods have been employed to train their methods. They have reported an accuracy of 85.48\% accuracy and 90.55\% sensitivity. This work has been extended by the same authors in [3]. In this paper they added new features in the form of CSF measurements and SPECT imaging markers. They reported an accuracy of 96.40\% and 97.03\% sensitivity. This paper has motived us to further the study. In the present paper an attempt has been made to improve the accuracy by using advanced machine learning models. Some recent machine learning algorithms have been chosen for prediction and have made a comparative performance analysis of these models based on accuracy, area under the ROC curve and other measures. 

\section{Materials and Methods}

A flowchart of the proposed analysis is shown in Fig 1. The data was first collected and the required non-motor and biomarker features are then extracted. Then different machine learning algorithms are employed for the classification task. Finally, a comparative analysis is made based on the accuracy provided by different machine learning models.

\subsection{Database}

In this study the data from Parkinson's Progression Markers Initiative (PPMI) database [8] was obtained. PPMI is an observational, multi‐centre study that collects clinical and imaging data and biologic samples from various cohorts that can be used by researchers to establish markers of disease progression in PD. PPMI has established a comprehensive, standardized, longitudinal PD data and biological sample repository that can play a vital role in the development of tools which assist in prediction of PD. To obtain the recent information, the official website of PPMI. ( “www.ppmi-info.org” ) can be visited. This dataset is similar to the one used in [3].  
We downloaded the database on 8th August 2016. On this date the data of 184 normal patients and 402 early PD subjects were collected. It is noted that PPMI has observations from each of the patients at different time intervals. Thus the data of each patient at different periods like screening or baseline, first visit, second visit and so on are available. In the present investigation the data at baseline observation are considered.

In [3] , the authors have used features from University of Pennsylvania Smell Identification Test, RBD screening questionnaire, CSF Markers of A\textbeta 1-42,\textalpha- syn, P-tau181, T-tau, T-tau/A\textbeta 1-42, P-tau181/A\textbeta 1-42 and P-tau181/T tau, and SPECT measurements of striatal binding ratio (SBR) data. In this study these features have been used because we felt that they are a good combination of non-motor features and biomarkers. The details of these features are given in section \rm III B.

\subsection{Feature Description}

\subsubsection{University of Pennsylvania Smell Identification Test (UPSIT)}

Olfactory dysfunction is an important marker of Parkinson's disease [9]. It acts as sensitive and early marker for Parkinson's disease. It is a fact that most of the people who suffer from PD have olfactory loss however it doesn`t mean that all the people with olfactory loss are suffering from PD [10]. Olfactory dysfunction are in various forms for instance it may be impairment in odour detection or odour differentiation. A study by Posen et al [11] showed that about 10\% of the subjects who were suffering from odour dysfunction were at the risk of PD.\\
For quantifying this odour loss the data of University of Pennsylvania Smell Identification Test is used. This test is commercially available and is also one of the most reliable tests [12]. The procedure of the test is as follows. A subject is provided with 4 different 10 page booklets. Each of this pages has a different odour. A subject has to scratch the page and smell it. For each of this pages, there exists a question with four options. Depending on the odour the subject selects one of the options. This procedure is repeated for all the pages in all the booklets. Once the test is completed the UPSIT score is calculated. The maximum score can be 40 when the subject identifies each of the odours correctly. One main advantage of this is that the test takes only a few minutes. For the present analysis the UPSIT score at baseline check-up from PPMI [8] has been taken. 

\subsubsection{REM sleep Behaviour Disorder Screening Questionnaire (RBDSQ)}

RBD is another non-motor symptom that plays an important role in early prediction of Parkinson's disease. People suffering from RBD have disturbances in sleep. These disturbances include vivid, aggressive or action packed dreams. Similar to olfactory loss, studies have shown that disorder in sleep behaviour increases the risk of being affected with Parkinson's disease.
For quantifying this non-motor symptom, the REM Sleep Behaviour Disorder Screening Questionnaire is used. The RBDSQ is a 10-item patient self-rating instrument [13]. The test contains ten short questions with answers as yes or no. A yes is equivalent to 1 and a no is equivalent to 0. The ten questions are divided such that each of the group of the questions provides  the observations about a particular behaviour.  Some of the examples of the questions from [13] are ``I sometimes have vivid dreams'', ``The dream contents mostly matches my nocturnal behaviour'', ``My sleep is frequently disturbed'', etc. As some of the subjects may have a bed partner, they can also be used in this test.\\
Each of the answers are provided as either one or zero. In the present study the feature for sleep disorder is obtained by summing up all the answers. This sum can be a maximum of 12 if we take the first nine questions. It is observed here that a higher score in this case means a higher risk of PD in contrast to that of UPSIT score. This RBDSQ score is taken from PPMI [8].

\begin{table*}[]
\centering
\caption{Performance Measures for various classifiers used in the study}
\label{Performance Measure}
\begin{tabular}{|l|l|l|l|l|l|l|l|l|}
\hline
\multirow{2}{*}{Performance Measures} & \multicolumn{2}{l|}{Multilayer Perceptron} & \multicolumn{2}{l|}{BayesNet} & \multicolumn{2}{l|}{Random Forest} & \multicolumn{2}{l|}{Boosted Logistic Regression} \\ \cline{2-9} 
                                      & Training             & Testing             & Training       & Testing      & Training         & Testing         & Training                & Testing                \\ \hline
Accuracy(\%)                          & 96.09                & 95.4545             & 96.5854        & 96.027       & 100              & 96.59           & 95.8537                 & 97.1591                \\ \hline
Recall                                & 0.961                & 0.955               & 0.966          & 0.960        & 1                & 0.966           & 0.959                   & 0.972                  \\ \hline
Precision                             & 0.962                & 0.955               & 0.967          & 0.965        & 1                & 0.970           & 0.959                   & 0.974                  \\ \hline
F-Measure                             & 0.961                & 0.955               & 0.966          & 0.961        & 1                & 0.967           & 0.959                   & 0.972                  \\ \hline
AUC                                   & 0.989                & 0.986               & 0.994          & 0.994        & 1                & 0.997           & 0.995                   & 0.989                  \\ \hline
\end{tabular}
\end{table*}

\subsubsection{Cerebrospinal Fluid Biomarkers}
Biomarkers play a pivotal role in this analysis. Without the aid of biomarkers the prediction of PD is less accurate. The biomarkers are the significant factors in increasing the accuracy of the model. Biomarkers need to be sensitive, reproducible and must be closely associated with the disease. 
Cerebrospinal fluid is a clear, colourless body fluid found in the brain. It has more physical contact with the brain as compared to any other fluid [14]. Due to the close proximity with the brain, any protein or peptide which is related to the brain specific functionalities or disease are diffused into CSF. Hence, the CSF can act as an important biomarker for brain related diseases which in the present case is Parkinson's disease. \\

The CSF samples are collected from PPMI. In PPMI, for each of subjects the CSF samples are obtained and certain measurements are made. These measurements include A\textbeta 1-42(amyloid beta (1-42), T-tau (total tau) and P-tau181 (tau phosphorylated at threonine) [15].  According to PPMI Research Laboratory these three are the important biomarkers that can be extracted from the CSF fluid. Along with this the concentration of \textalpha -Syn was also collected from PPMI database. 
Kang et al have mentioned that ratios like T- tau/A\textbeta 1-42, P-tau181/A\textbeta 1-42 and P-tau181/T-tau also play a significant role in early detection of Parkinson's disease [16]. In the present investigation the measurements of A\textbeta 1-42, T-tau and P-tau181 and also the ratios T- tau/A\textbeta 1-42, P-tau181/A\textbeta 1-42 and P-tau181/T-tau are taken.

\subsubsection{Neuroimaging markers}
Single-photon emission computed tomography (SPECT) is a neuroimaging technique that uses gamma rays [17]. The SPECT is a common routine for helping a doctor to decide whether a subject is suffering from neurodegenerative diseases. According to [18], the SPECT imaging can detect the dopaminergic transporter loss during the early stages of PD.  When a subject has an abnormal scanning then the person has more probability of being affected with Parkinson's disease or other neuro degenerative disease. However, a normal scan denotes that the subject is suffering from other type of diseases [18].\\
DatScan SPECT imaging obtained from PPMI imaging centres are used in this study. At PPMI the striatal binding ratios were calculated. The DatScan SPECT images are collected according to the PPMI imaging protocol. This raw images are then reconstructed so as to ensure consistency among different imaging centres. After this attenuation correction is performed on these images. After this the Gaussian filter is applied and it is followed by normalization. Finally the required part is extracted from the images and then the striatal binding ratio for left and right caudate, the left and right putamen  are calculated [19]. In this paper, these four striatal binding values are used as neuroimaging biomarkers.

\subsection{Prediction models for distinguishing early PD and healthy normal subjects}

In this study, four different machine learning classifiers are chosen for classification task. A brief description of each of them is provided in this section. WEKA [20] is used for classification using Multilayer Perceptron, Bayesian Network, Random Forest, and Boosted Logistic Regression. The main motive is to find an algorithm that can improve the already reported accuracy as well as to see how various models are performing. Firstly, the dataset is normalized using the Normalize filter in WEKA [20].Then then the dataset is divided in such a way that 70\% is used for training and the rest 30\% is used for testing. While partitioning the dataset the same class proportion in both the test and train data is maintained. For example, if the proportion of healthy people in the complete data is 40\% then both in training and testing the proportion of healthy people to PD subjects is maintained at 40\%. This type of partitioning is known as stratified partitioning. The accuracy, recall, precision and f-measure for each these algorithms are computed and the ROC of each of the classifiers are plotted. Finally the performance measure of different classifiers used in this paper as well as in [3] are compared. 

\subsubsection{Multilayer Perceptron}
Multilayer perceptron is a feedforward artificial neural network. The basic principle of multilayer perceptron is that it takes the input and maps it to a nonlinear space, then it tries to predict the corresponding outputs. A MLP architecture is viewed as a multiple layers of nodes, with each layer being fully connected with the next layer. Each node in the MLP is interpreted as a neuron that has an activation function which is non-linear [21] [22]. The back-propagation algorithm which is a supervised learning technique is used for training the model.  The number of hidden layers in the MLP have a significant impact on the performance of the classifier.

\subsubsection{Bayesian Network}
The Bayesian network is one of the probabilistic graphical models used in machine learning. The Bayes Net corresponds to graphical model structures which are known as directed acyclic graph (DAG). This graphical models are understood in the following manner [23]. The nodes in the graph represent the random variables and the edge between node x and node y denotes the probabilistic dependencies among random variables corresponding to the respective nodes. Hence the nodes that are not connected in the Bayesian network are the random variables which are independent to each other. Different computational and statistical methods are used to estimate the conditional dependencies.  Bayes Network learning uses various search algorithms and quality measures. In the present model K2 learning algorithm for searching is used.

\subsubsection{Random Forest}
Random forest are part of ensemble learning method that is used for classification, regression and other tasks. In Random forest, there are many decision trees. For a given input, each of the decision trees classify it as yes/no (in case of binary classification)[24] [25]. Then once each of the trees have classified as yes/no, the value which has the majority among them is taken as output. The advantages are that this algorithm runs effectively on large inputs and it also helps in estimating which of the features are important.

\subsubsection{Boosted Logistic Regression}
Logistic regression was developed by statistician David Cox in 1958[26] [27]. A logistic model is used to predict the binary class using one or more features. Logit- the natural algorithm for an odds ratio is the central mathematical concept behind logistic regression. Logistic regression is well suited in case when one wants to establish relationship between a categorical outcome variable and one or more categorical or continuous predictor variables [28]. 

Boosting is a machine learning ensemble meta-algorithm for primarily reducing bias, and also variance in supervised learning. It belongs to the family of machine learning algorithms which convert weak learners to strong ones. AdaBoost is used for boosting different classifiers. 
\begin{table*}[]
\centering
\caption{Comparative Analysis of Machine Learning models \\ in the current work and previous work }
\label{my-label}
\begin{tabular}{|l|l|l|l|l|}
\hline
\multirow{2}{*}{Machine Learning Algorithms} & \multicolumn{2}{l|}{Accuracy(\%)} & \multicolumn{2}{l|}{AUC(\%)} \\ \cline{2-5} 
                                             & Training         & Testing        & Training      & Testing      \\ \hline
Multilayer Perceptron                        & 96.09            & 95.45          & 98.9          & 98.6         \\ \hline
BayesNet                                     & 96.5854          & 96.02          & 99.4          & 99.4         \\ \hline
Random Forest                                & 100              & 96.59          & 100           & 0.997        \\ \hline
Boosted Logistic Regression                  & 95.8537          & 97.16          & 99.5          & 98.9         \\ \hline
Boosted Trees                                & 100              & 95.08          & 100           & 98.23        \\ \hline
Naive Bayes                                  & 94.67            & 93.12          & 98.66         & 96.77        \\ \hline
Support Vector Machine                       & 97.14            & 96.40          & 99.27         & 98.88        \\ \hline
Logistic Regression                          & 96.50            & 95.63          & 99.20         & 98.66        \\ \hline
\end{tabular}
\end{table*}

\begin{figure*}[t]
\begin{subfigure}{.5\textwidth}
  \centering
  \fbox{\includegraphics[width=.8\linewidth ,height = 5cm]{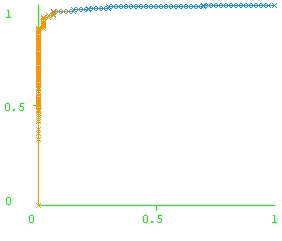}}
  \caption{ROC for Classification using Multilayer Perceptron (Test Data)}
  \label{fig: fig:ROC for Classification using Multilayer Perceptron (Test Data)}
\end{subfigure}%
\begin{subfigure}{.5\textwidth}
  \centering
  \fbox{\includegraphics[width=.8\linewidth , height = 5cm]{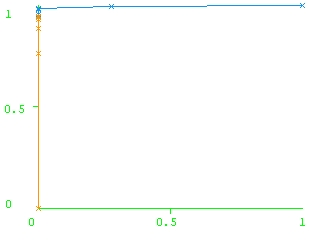}}
  \caption{ROC for Classification using BayesNet (Test Data)}
  \label{fig: ROC for Classification using BayesNet (Test Data)}
\end{subfigure}

\begin{subfigure}{.5\textwidth }
  \centering
  \fbox{\includegraphics[width=.8\linewidth , height = 5cm]{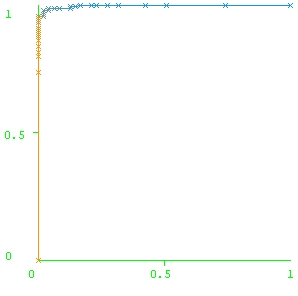}}
  \caption{1f ROC for Classification using Random Forest (Test Data)}
  \label{fig: ROC for Classification using Random Forest (Test Data)}
\end{subfigure}%
\begin{subfigure}{.5\textwidth}
  \centering
  \fbox{\includegraphics[width=.8\linewidth, height = 5cm]{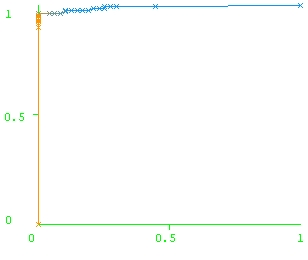}}
  \caption{ROC for Classification using Boosted Logistic Regression
  \\(Test Data)}
  \label{fig: ROC for Classification using Boosted Logistic Regression(Test Data)}
\end{subfigure}
\caption{ROC Plots for different machine learning algorithms
\\X-axis = False Positive Rate Y-axis = True Positive Rate}
\label{fig:ROC Curves}
\end{figure*}

\section{Results and Discussion}

Table 1 shows the performance of various classifiers used in the study. Fig 2 shows the corresponding ROC plots. In Multilayer Perceptron the back propagation algorithm is used to train the model. The learning rate is set at 0.4 and number of hidden layers is chosen as 8 in this case ((number of attributes + number of classes)/2 =8). In BayesNet various search algorithms and quality measures are used. A Simple Estimator is chosen for estimating the conditional probability tables of a Bayes network once the structure has been learned. For searching K2 algorithm is used. It uses a hill climbing algorithm which is restricted by an order on the variables. In boosted logistic regression Adaboost M1 method is used to boost the logistic regression.\\ 

It is observed that all the classifiers performed reasonably well with boosted logistic regression giving the best performance with 97.16\% accuracy and 98.9\% area under the ROC (AUC).  Table 2: shows how this models performed in relation to the previous work [3]. It is found that the accuracy and area under the ROC curve are nearly same among the different classifiers used. The present work and [3] have the advantage that the dataset used is very large as compared to others. However, it is noted that the PPMI study includes subjects who are in early stages of PD and healthy normal, however it doesn`t include subjects who are having premotor symptoms but are not diagnosed as PD due to lack of  motor symptoms.

\section{Conclusion}

The diagnosis of Parkinson's Disease is not direct which means that one particular test like blood test or ECG cannot determine whether a person is suffering from PD or not. Doctors go through the medical history of a patient, followed by a thorough neurological examination. They find out at least two cardinal symptoms among the subjects and then predict whether the subject is suffering from PD. The misdiagnosis rate of PD is significant due to a no definitive test. In such case it will be helpful for us to aid the doctor by providing a machine learning model. The prediction models are developed using machine learning techniques of boosted logistic regression, classification trees , Bayes Net and multilayer perceptron based on these significant features. It is observed that the performance is better. It is demonstrated that Boosted Logistic Regression produce superior results. These results encourage us to try other ensemble learning techniques. The present work employs different machine learning algorithms which are not used in [3]. This study plays an important role in having a comparative analysis of various machine learning algorithms. 
In conclusion, this models can provide the nuclear experts an assistance that can aid them in better and accurate decision making and clinical diagnosis. It is also found that the proposed method is fully automated and provides improved performance and hence can be recommended for real life applications.


\begin{thebibliography}{}
\bibitem{1}
Alves G, Forsaa EB, Pedersen KF, Dreetz Gjerstad M, Larsen JP (2008) Epidemiology of Parkinson's disease. J Neurol 255 Suppl 5: 18–32.
\bibitem{2}
Vu TC, Nutt JG, Holford NH (2012) Progression of Motor and Non-Motor Features of Parkinson's Disease and Their Response to Treatment. Br J Clin Pharmacol. 
\bibitem{3}
High Accuracy Detection of Early Parkinson's Disease through Multimodal Features and Machine Learning. R Prashant, Sumantra Dutta Roy, Pravat K. Mandal, Shantanu Ghosh.
\bibitem{4}
Rustempasic, Indira, \& Can, M. (2013). Diagnosis of Parkinson's disease using Fuzzy C-Means Clustering and Pattern Recognition. SouthEast Europe Journal of Soft Computing, 2(1).
\bibitem{5}
Amit S., Ashutosh M., A. Bhattacharya, F. Revilla, (2014, March).Understanding Postural Response of Parkinson's Subjects Using Nonlinear Dynamics and Support Vector Machines. Austin J, Biomed Eng 1(1): id1005.
\bibitem{6}
L. Silveira-Moriyama, J. Carvalho Mde, R. Katzenschlager, A. Petrie, R. Ranvaud, E.R. Barbosa, A.J. Lees, The use of smell identification tests in the diagnosis of Parkinson's disease in Brazil, Mov Disord, 23 (2008) 2328-2334. 
\bibitem{7}
Parkinson's disease detection using olfactory loss and REM sleep disorder features. R. Prashanth-EMBS Member, S. Dutta Roy-IEEE Member, and P. K. Mandal, S. Ghosh 
\bibitem{8}
K. Marek, D. Jennings, S. Lasch, A. Siderowf, C. Tanner, T. Simuni, C. Coffey, K. Kieburtz, E. Flagg, S. Chowdhury, W. Poewe, B. Mollenhauer, P.-E. Klinik, T. Sherer, M. Frasier, C. Meunier, A. Rudolph, C. Casaceli, J. Seibyl, S. Mendick, N. Schuff, Y. Zhang, A. Toga, K. Crawford, A. Ansbach, P. De Blasio, M. Piovella, J. Trojanowski, L. Shaw, A. Singleton, K. Hawkins, J. Eberling, D. Brooks, D. Russell, L. Leary, S. Factor, B. Sommerfeld, P. Hogarth, E. Pighetti, K. Williams, D. Standaert, S. Guthrie, R. Hauser, H. Delgado, J. Jankovic, C. Hunter, M. Stern, B. Tran, J. Leverenz, M. Baca, S. Frank, C.-A. Thomas, I. Richard, C. Deeley, L. Rees, F. Sprenger, E. Lang, H. Shill, S. Obradov, H. Fernandez, A. Winters, D. Berg, K. Gauss, D. Galasko, D. Fontaine, Z. Mari, M. Gerstenhaber, D. Brooks, S. Malloy, P. Barone, K. Longo, T. Comery, B. Ravina, I. Grachev, K. Gallagher, M. Collins, K.L. Widnell, S. Ostrowizki, P. Fontoura, T. Ho, J. Luthman, M.v.d. Brug, A.D. Reith, P. Taylor, The Parkinson Progression Marker Initiative (PPMI), Prog Neurobiol, 95 (2011) 629-635.
\bibitem{9}
Olfactory dysfunction in Parkinson disease Richard L. Doty
\bibitem{10}
https://www.michaeljfox.org/understanding-parkinsons/living-with-pd/topic.php?smell-loss
\bibitem{11}
M.M. Ponsen, D. Stoffers, J. Booij, B.L. van Eck-Smit, E.C. Wolters, H.W. Berendse, Idiopathic hyposmia as a preclinical sign of Parkinson's disease, Ann Neurol, 56 (2004) 173-181.
\bibitem{12}
R.L. Doty, P. Shaman, M. Dann, Development of the University of Pennsylvania Smell Identification Test: a standardized microencapsulated test of olfactory function, Physiol Behav, 32 (1984) 489-502
\bibitem{13}
The REM Sleep Behaviour Disorder Screening Questionnaire – A new Diagonostic Instrument, Karin Stiasny-Kolster, Geert Mayer, Hephata Klinik Schwalmstadt-Treysa, Germany,Sylvia Schäfer, Wolfgang H Oertel
\bibitem{14}
Cerebrospinal fluid biomarkers in parkinsonian conditions: an update and future directions
Nadia Magdalinou, Andrew J Lees, Henrik Zetterberg.
\bibitem{15}
A\textbeta 1-42, t-tau and p-tau181 measurements in 120 PPMI CSF samples using xMAP/Luminex multiplex immunoassay, Ju Hee Kang, John Q Trojanowski and Leslie M Shaw, Department of Pathology \& Laboratory Medicine and Institute onAging, Center for Neurodegenerative Disease Research, Perelman School of Medicine, University of Pennsylvania
\bibitem{16}
J.H. Kang, D.J. Irwin, A.S. Chen-Plotkin, A. Siderowf, C. Caspell, C.S. Coffey, T. Waligórska, P. Taylor, S. Pan, M. Frasier, K. Marek, K. Kieburtz, D. Jennings, T. Simuni, C.M. Tanner, A. Singleton, A.W. Toga, S. Chowdhury, B. Mollenhauer, J.Q. Trojanowski, L.M. Shaw, Parkinson's Progression Markers Initiative, Association of cerebrospinal fluid β-amyloid 1-42, T-tau, P-tau181, and α-synuclein levels with clinical features of drug-naive patients with early Parkinson disease, JAMA Neurol, 70 (2013) 1277-1287.
\bibitem{17}
SPECT at the US National Library of Medicine Medical Subject Headings (MeSH)
\bibitem{18}
J.L. Cummings, C. Henchcliffe, S. Schaier, T. Simuni, A. Waxman, P. Kemp, The role of dopaminergic imaging in patients with symptoms of dopaminergic system neurodegeneration, Brain, 134 (2011) 3146-3166.
\bibitem{19}
DatScan SPECT Image Processing Methods for Calculation of Striatal Binding Ratio (SBR), Gary Wisniewski, John Seibyl, Ken Marek, Institute for Neurodegenerative Disorders (IND)
\bibitem{20}
Mark Hall, Eibe Frank, Geoffrey Holmes, Bernhard Pfahringer, Peter Reutemann, Ian H. Witten (2009); The WEKA Data Mining Software: An Update; SIGKDD Explorations, Volume 11, Issue 1.
\bibitem{21}
Rosenblatt, Frank. x. Principles of Neurodynamics: Perceptrons and the Theory of Brain Mechanisms. Spartan Books, Washington DC, 1961
\bibitem{22}
Rumelhart, David E., Geoffrey E. Hinton, and R. J. Williams. "Learning Internal Representations by Error Propagation". David E. Rumelhart, James L. McClelland, and the PDP research group. (editors), Parallel distributed processing: Explorations in the microstructure of cognition, Volume 1: Foundations. MIT Press, 1986.
\bibitem{23}
Ben-Gal I., Bayesian Networks, in Ruggeri F., Faltin F. \& Kenett R.,Encyclopedia of Statistics in Quality \& Reliability, Wiley \& Sons (2007).
\bibitem{24}
Ho, Tin Kam (1995). Random Decision Forests (PDF). Proceedings of the 3rd International Conference on Document Analysis and Recognition, Montreal, QC, 14–16 August 1995. pp. 278–282.
\bibitem{25}
Ho, Tin Kam (1998). "The Random Subspace Method for Constructing Decision Forests" (PDF). IEEE Transactions on Pattern Analysis and Machine Intelligence. 20 (8): 832–844. doi:10.1109/34.709601
\bibitem{26}
Walker, SH; Duncan, DB (1967). "Estimation of the probability of an event as a function of several independent variables". Biometrika. 54: 167–178. doi:10.2307/2333860
\bibitem{27}
Cox, DR (1958). "The regression analysis of binary sequences (with discussion)". J Roy Stat Soc B. 20: 215–242.
\bibitem{28}
An Introduction to Logistic Regression Analysis and Reporting CHAO-YING JOANNE PENG KUK LIDA LEE GARY M. INGERSOLL Indiana University-Bloomington

\end{thebibliography}
\end{document}